\documentclass[10pt,twocolumn,letterpaper]{article}

\usepackage{cvpr}
\usepackage{times}
\usepackage{epsfig}
\usepackage{graphicx}
\usepackage{amsmath}
\usepackage{amssymb}

\usepackage[bookmarks=false]{hyperref}

\cvprfinalcopy 

\def\cvprPaperID{****} 

\begin{document}

\title{Exploring Context, Attention and Audio Features\\ for Audio Visual Scene-Aware Dialog}

\author{First Author\\
Institution1\\
Institution1 address\\
{\tt\small firstauthor@i1.org}
\and
Second Author\\
Institution2\\
First line of institution2 address\\
{\tt\small secondauthor@i2.org}
}
\author{Shachi H Kumar
 \qquad 
Eda Okur
 \qquad 
Saurav Sahay
 \qquad 
Jonathan Huang
 \qquad 
Lama Nachman\\
Intel Labs, Anticipatory Computing Lab, USA\\
{\tt\small \{shachi.h.kumar, eda.okur, saurav.sahay, jonathan.huang, lama.nachman\}@intel.com}
}

\maketitle



\section{Introduction}
We are witnessing a confluence of vision, speech and dialog system technologies that are enabling the IVAs to learn audio-visual groundings of utterances and have conversations with users about the objects, activities and events surrounding them. Recent progress in visual grounding techniques \cite{7410636, Das_2017} and Audio Understanding \cite{hershey2017cnn} are enabling machines to understand shared semantic concepts and listen to the various sensory events in the environment. With audio and visual grounding methods~\cite{yu2016video, Hori_2017}, end-to-end multimodal SDS~\cite{serban2016building} are trained to meaningfully communicate with us in natural language about the real dynamic audio-visual sensory world around us. In this work, we explore the role of `topics' as the context of the conversation along with multimodal attention into such an end-to-end audio-visual scene-aware dialog system architecture. We also incorporate an end-to-end audio classification ConvNet, AclNet, into our models.  We develop and test our approaches on the Audio Visual Scene-Aware Dialog (AVSD) dataset \cite{DBLP:journals/corr/abs-1806-00525} released as a part of the DSTC7. We present the analysis of our experiments and show that some of our model variations outperform the baseline system~\cite{DBLP:journals/corr/abs-1806-08409} released for AVSD.

\begin{figure*}
\centering
   \includegraphics[width=\textwidth]{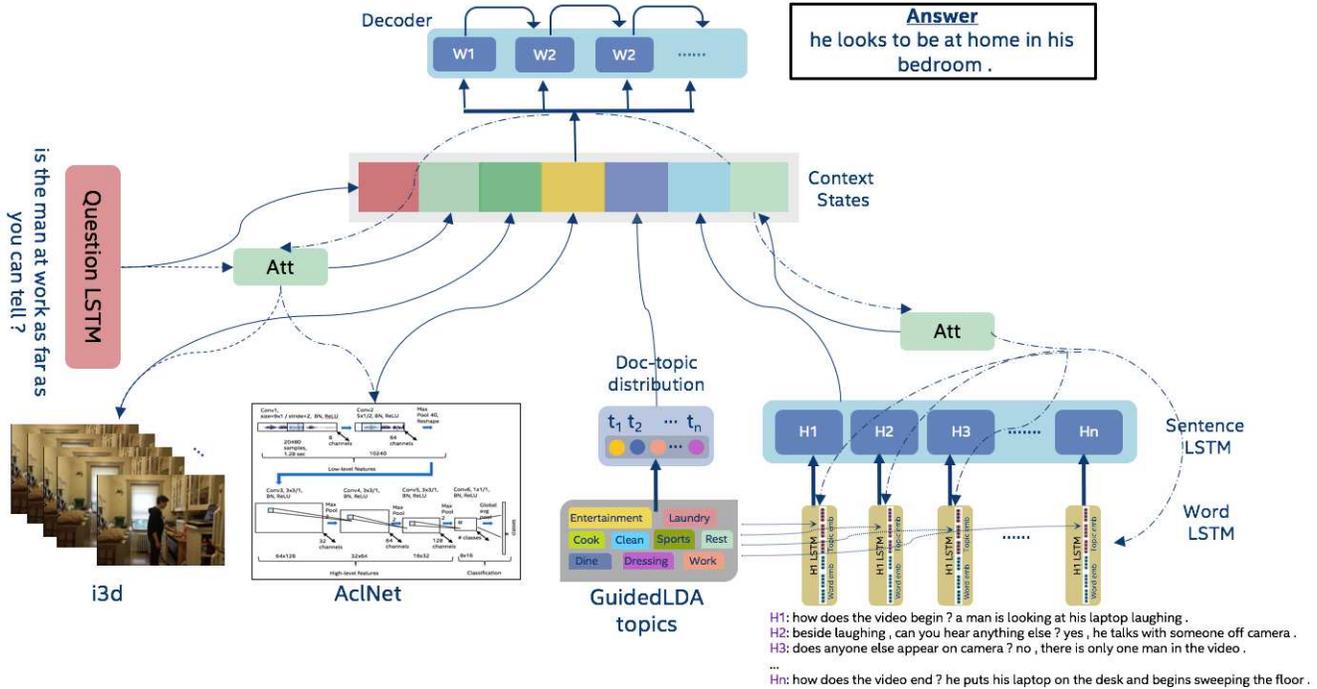}
   \caption{System Architecture}
\label{arch}
\end{figure*}

\section{Model Description}
In this section, we describe the main architecture explorations of our work\footnote{Further details can be found in the full-paper version of this work~\cite{kumar2019context}.} as shown in Figure \ref{arch}. 

\textbf{{Topic Model Explorations}}: 
Topics form a very important source of context in a dialog. 
We train Latent Dirichlet Allocation (LDA~\cite{blei2003latent}) and Guided LDA 
\cite{DBLP:conf/eacl/JagarlamudiDU12} models on questions, answers, QA pairs, captions and history and incorporate the topic distributions as features or use them to learn topic embeddings. Since we are interested in identifying specific topics (e.g., entertainment, cooking, cleaning), we use Guided LDA to generate topics based on seedwords.

\textbf{{Attention Explorations}}:  
 We explore several configurations of the model where at every step, the decoder attends to the dialog history representations and AV features to selectively focus on relevant parts of the dialog history and audio/video. This helps create a combination of the dialog history and multimodal context that is richer than the single context vectors of the individual modalities.
 
\textbf{Audio Feature Explorations}: We used an end-to-end audio classification ConvNet, called AclNet \cite{AclNet}. AclNet takes raw, amplitude-normalized 44.1 kHz audio samples as input, and produces classification output without the need to compute spectral features. AclNet is trained using the ESC-50 \cite{piczak2015dataset} corpus, a dataset of 50 classes of environmental sounds organized in 5 semantic categories (animals, interior/domestic, exterior/urban, human, natural landscapes).  

\section{Dataset}
We use the dialog dataset~\cite{DBLP:journals/corr/abs-1806-00525} consisting of conversations between two people about a video (from Charades human action dataset~\cite{DBLP:conf/eccv/SigurdssonVWFLG16}), which was released as part of the AVSD challenge at DSTC7. We use the official-training (7659 dialogs) and prototype-validation sets (732 dialogs) to train, and prototype-test set (733 dialogs) to evaluate our models. 

\section{Experiments and Results}

\begin{table}[b]
\scriptsize
\begin{center}
    \resizebox{\columnwidth}{!}{
    \begin{tabular}{| l | l | l | l | l | l | l | l |}
    \hline
      & Bleu1 &  Bleu2 &  Bleu3 &  Bleu4 & Meteor & Rouge & CIDEr \\
    \hline
    Baseline & 
    0.273 & 0.173 & 0.118 & 0.084 & 0.117 & 0.291 & 0.766 \\
    StandardLDA & 
    0.255 & 0.164 & 0.113 & 0.082 & 0.114 & 0.285 & \textbf{0.772} \\
    GuidedLDA & 
    0.265 & 0.170 & 0.117 & 0.084 & \textbf{0.118} & \textbf{0.293} & \textbf{0.812} \\
    GuidedLDA-all & 
    0.272 & 0.173 & 0.118 & \textbf{0.085} & \textbf{0.119} & \textbf{0.293} & \textbf{0.793} \\
    GuidedLDA+GloVe & 
    \textbf{0.275} & \textbf{0.175} & \textbf{0.119} & \textbf{0.085} & \textbf{0.121} & \textbf{0.293} & \textbf{0.797} \\
    Topic-embeddings & 
    0.257 & 0.165 & 0.114 & 0.083 & 0.115 & 0.287 & \textbf{0.772} \\
    HLSTM-with-topics & 
    0.260 & 0.166 & 0.115 & 0.084 & 0.117 & 0.290 & \textbf{0.797} \\
    \hline
    \end{tabular}
    }
    \caption{Topic Experiments}
    \label{table:all-topics}
\end{center}
\end{table}

{\textbf{Topic Experiments}}: We use separate topic models trained on questions, answers, QA pairs, captions, history and history+captions to generate topics for samples from each category. Table~\ref{table:all-topics} compares the baseline model with the addition of StandardLDA and GuidedLDA topic distributions as features to the decoder, as well as by learning topic embeddings. In general, GuidedLDA models perform better than StandardLDA, and GuidedLDA + GloVe~\cite{pennington2014glove} is our best performing model.   

 \begin{table}[t]
 \scriptsize
\begin{center}
  \resizebox{\columnwidth}{!}{
  \begin{tabular}{| l | l | l | l | l | l | l | l |}
    \hline
 &\scriptsize Bleu1& \scriptsize	Bleu2& \scriptsize	Bleu3& \scriptsize	Bleu4& \scriptsize	Meteor& \scriptsize	Rouge& \scriptsize	CIDEr \\ 
 \hline
 \scriptsize   \textbf{Overall} & & & & & & & \\
 \scriptsize   Baseline(B)& \scriptsize	0.273 & \scriptsize	0.173 & \scriptsize	0.118& \scriptsize	0.084& \scriptsize	0.117& \scriptsize	0.291& \scriptsize	0.766 \\ 
\scriptsize    B+VGGish& \scriptsize	0.271& \scriptsize	0.172& \scriptsize	0.118	& \scriptsize0.085& \scriptsize	0.116& \scriptsize	0.292& \scriptsize	\textbf{0.791}\\ 
\scriptsize    B+AclNet& \scriptsize	\textbf{0.274}& \scriptsize	\textbf{0.175}& \scriptsize	\textbf{0.121}& \scriptsize	\textbf{0.087}& \scriptsize	0.117& \scriptsize	\textbf{0.294}& \scriptsize	0.789 \\    
\hline

\scriptsize   \textbf{Audio-related} & & & & & & & \\
\scriptsize Baseline(B)& \scriptsize \textbf{0.267}& \scriptsize 0.179& \scriptsize 0.128& \scriptsize 0.096& \scriptsize 0.120& \scriptsize 0.285& \scriptsize 0.919 \\ 
\scriptsize B+VGGish& \scriptsize 0.266& \scriptsize 0.181& \scriptsize 0.131& \scriptsize 0.099& \scriptsize 0.118& \scriptsize 0.285& \scriptsize 0.907
\\ 
\scriptsize B+AclNet& \scriptsize 0.266& \scriptsize \textbf{0.183}& \scriptsize \textbf{0.132}& \scriptsize \textbf{0.100}& \scriptsize 0.120& \scriptsize \textbf{0.287}& \scriptsize \textbf{0.944} \\ 
\hline
  \end{tabular}
  }
    \caption{Audio Feature Experiments}
    \label{table:audioexp_final}
\end{center}
\end{table}

\textbf{{Audio Experiments}}: Table~\ref{table:audioexp_final} shows the comparison of the baseline B (without audio features), and B+VGGish (provided as a part of the AVSD task) and B+AclNet features. We analyse the effects of audio features on the overall dataset as well as specifically on audio-related dialogs. From Table ~\ref{table:audioexp_final}, we observe that B+AclNet performs the best both on overall dataset and audio-related dialogs.

\begin{figure}
   \includegraphics[width=\columnwidth]{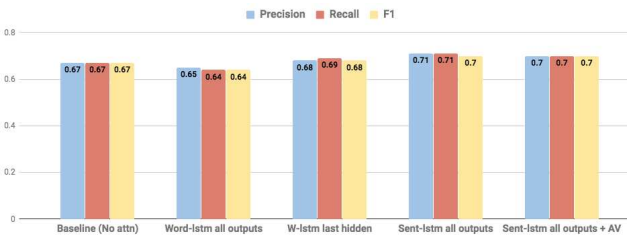}
   \caption{Precision, Recall, F1 Scores for Attention Experiments on Questions Containing Coreferences}
\label{attentionCoref}
\end{figure}

\textbf{{Attention Experiments}}: Table \ref{table:attn} shows that the configuration where decoder attends to all of the sentence-LSTM output states performs better than the baseline. In order to compare the results based on semantic meaningfulness, we performed quantitative analysis on dialogs that contained binary answer in the ground truth. We evaluate our models on their ability to predict these binary answers correctly and present this analysis in Figure \ref{attentionCoref} which shows once again that the configuration where decoder attends to all of the sentence-LSTM output states performs best on binary answer evaluation.

 \begin{table}[t]
\begin{center}
\resizebox{\columnwidth}{!}{
\scriptsize
  \begin{tabular}{| l | l | l | l | l | l | l | l |}
    \hline
 &\scriptsize Bleu1& \scriptsize	Bleu2& \scriptsize	Bleu3& \scriptsize	Bleu4& \scriptsize	Meteor& \scriptsize	Rouge& \scriptsize	CIDEr \\ 
 \hline
 
 \scriptsize Baseline	& \scriptsize \textbf{0.248}	& \scriptsize 0.151& \scriptsize	0.101& \scriptsize	0.071& \scriptsize 0.110 & \scriptsize 0.256 & \scriptsize 0.664\\ 
 
  \scriptsize Word-LSTM all	& \scriptsize 0.223	& \scriptsize 0.138& \scriptsize	0.092& \scriptsize	0.065& \scriptsize
 0.103& \scriptsize	0.262& \scriptsize	0.591  \\ 

 \scriptsize W-LSTM last & \scriptsize	0.229& \scriptsize	0.139& \scriptsize	0.093& \scriptsize	0.065& \scriptsize	0.105& \scriptsize	{0.250}& \scriptsize	{0.661} \\ 

 \scriptsize Sent-LSTM all & \scriptsize 	0.242& \scriptsize	{0.151}& \scriptsize	\textbf{0.103}& \scriptsize	\textbf{0.073}& \scriptsize	{0.110}& \scriptsize	\textbf{0.261}& \scriptsize	\textbf{0.707 }\\ 
 
  \scriptsize S-LSTM all + AV& \scriptsize 	0.234& \scriptsize	{0.146}& \scriptsize	{0.099}& \scriptsize	{0.070}& \scriptsize	{0.109}& \scriptsize	{0.254}& \scriptsize	\textbf{{0.690 }}\\ 
 
 \hline

  \end{tabular}
  }
    \caption{Decoder Attention over Dialog History and AV Features}    
    \label{table:attn}
\end{center}
\end{table}

\section{Conclusion}
In this paper, we present some explorations and techniques for contextual and multimodal end-to-end audio-visual scene aware dialog system. We incorporate context of the dialog in the form of topics, we use various attention mechanisms to enable the decoder to focus on relevant parts of the dialog history and audio/video features, and incorporate audio features from an end-to-end audio classification architecture, AclNet. We validate our approaches on the AVSD dataset and show that these techniques give better performance compared to the baseline.

{\small
\bibliographystyle{ieee_fullname}
\bibliography{cvpr_2page_arXiv}
}

\end{document}